\documentclass[openacc]{rstransa}
\usepackage[caption=false,font=footnotesize]{subfig}
\usepackage{algorithm} 
\usepackage{algorithmicx}
\usepackage{algpseudocode}
\usepackage[flushleft]{threeparttable} 
\usepackage{multirow} 
\usepackage{array}
\usepackage{boldline}
\usepackage{graphicx} 
\usepackage{hyperref}
\usepackage{multibib}
\usepackage[left]{lineno} 
\usepackage{float}

\newcommand{\ynote}[1]{{#1}} 
\newcommand{\newtext}[1]{{#1}}

\titlehead{Research}

\begin{document}


\title{Ensemble of Weak Spectral Total Variation Learners: a PET-CT Case Study}

\author{
Anna Rosenberg$^{1}$, John Kennedy$^{2}$, Zohar Keidar$^{3}$, Yehoshua Y. Zeevi$^{4}$ and Guy Gilboa$^{5}$}

\address{$^{1}$srosnbrg@campus.technion.ac.il\\
$^{2}$j\_kennedy@rambam.health.gov.il\\
$^{3}$z\_keidar@rambam.health.gov.il\\
$^{4}$zeevi@ee.technion.ac.il\\
$^{5}$guy.gilboa@ee.technion.ac.il}

\subject{xxxxx, xxxxx, xxxx}


\corres{Guy Gilboa\\
\email{guy.gilboa@ee.technion.ac.il}}

\begin{abstract}
Solving computer vision problems through machine learning, one often encounters lack of sufficient training data. 
To mitigate this we propose the use of ensembles of weak learners based on 
spectral total-variation (STV) features (Gilboa 2014). 
The features are related to nonlinear eigenfunctions of the total-variation subgradient and can characterize well textures at various scales. It was shown (Burger et-al 2016) that, in the one-dimensional case, orthogonal features are generated, whereas in two-dimensions the features are empirically lowly correlated.
Ensemble learning theory advocates the use of lowly correlated weak learners. We thus propose here to design ensembles using learners based on STV features.

To show the effectiveness of this paradigm we examine a hard real-world medical imaging problem:
the predictive value of computed tomography (CT) data for high uptake in positron emission tomography (PET) for patients suspected of skeletal metastases. 
The database consists of 457 scans
with 1524 unique 
 pairs of registered CT and PET slices. 
Our approach is compared to deep-learning methods and to 
Radiomics features, showing
STV learners perform best (AUC=$0.87$), compared to neural nets 
(AUC=$0.75$) and Radiomics (AUC=$0.79$). 
We observe that fine STV scales in CT images are especially indicative for the presence of high uptake in PET.
\end{abstract}


\begin{fmtext}

\end{fmtext}


\maketitle

\section{Introduction}
\label{sec:introduction}
Machine learning is highly instrumental in solving various computer vision tasks.
Exceptional results are obtained when abundance of data exists for training. In some domains, however, labeled data for training is scarce, such as in detecting specific diseases based on medical imaging scans. A common practice to overcome data scarcity in (artificial) neural networks (NN's), is to use pre-trained models and transfer-learning approaches. However, the performance of such techniques is not always sufficient. 
Part of the problem is that the extracted features are of low relevance for the specific task at hand.  

\begin{figure}[ht]
\centering
\subfloat{%
       \includegraphics[width=3in]{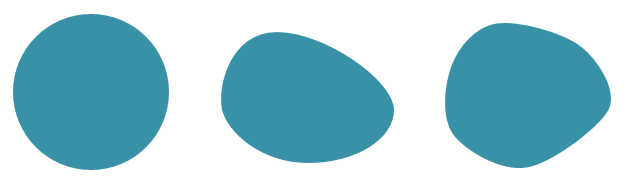}}
\caption{Examples of TV eigenfunctions. In $\mathbb{R}^2$, indicator functions of convex sets, where the maximal curvature is below the area-perimeter ratio, are eigenfunctions with respect to the subdifferential of TV, admitting the nonlinear eigenvalue problem $\lambda u \in \partial J_{TV}(u)$. Spectral TV \cite{gilboa_siam} approximates a decomposition of the signal into this type of structure at various scales.}
\label{fig:tv_efs}
\end{figure}

\begin{figure}[ht]
\centering 
\includegraphics[width=3.49in]{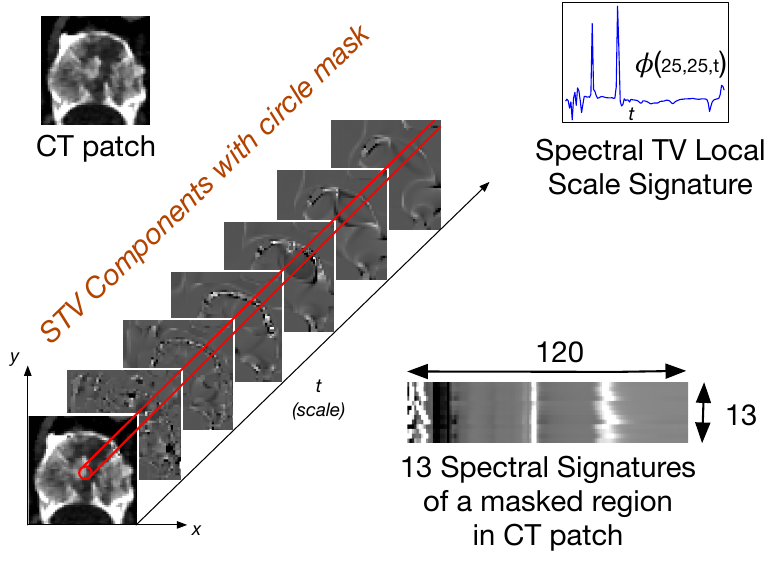}
\caption{Feature Extraction Overview. Spectral TV transform is applied on a CT patch and 120 Spectral components are obtained. Each pixel at location $(x,y)$ is described by a Spectral TV Local scale signature, $\phi(x,y,t)$. The signatures of 13 pixels that are inside the \ynote{circular} mask are used as features.}
\label{fig:feat_ext}
\end{figure}

In this study we aim to examine the usability of features related to the total variation functional in the context of medical imaging, and specifically, for \emph{computed tomography} (CT) analysis.
Total variation (TV) has shown to be a high-quality regularizer in medical imaging applications, since it is edge preserving and promotes low-curvature shapes. To analyze features related to TV one can use the TV transform, or spectral TV   \cite{gilboa_springer,gilboa_ssvm,gilboa_siam}.
To extract image features a (sub)-gradient flow with respect to the TV functional is evolved, initialized with the desired image. A spectral band is extracted by applying a second order numerical time-derivative, where time represents the spectral scale. 
It was shown that when the initial image is a solution of a nonlinear eigenvalue problem with respect to the TV subgradient (see examples in Fig. \ref{fig:tv_efs}), a single response occurs at a time inversely proportional to the eigenvalue. In the general case, the image is decomposed into blob-like features at increasing scale, in an edge-preserving manner, as shown in Fig. \ref{fig:feat_ext}.
A comprehensive theory for this representation was established in \cite{burger}. A significant result is that the spectral responses at all scales are orthogonal to each other, in one-dimension, for any initial condition. Theoretical underpinning in the
continuous case was further investigated in 
\cite{burger}.

Our aim here is to design a robust classifier, based on STV features. We use an ensemble of weak classifiers, each based on different spectral bands. The low correlation of these features yields a very robust design. 
The classifier is applied to solve a hard real-world medical imaging problem. 
We attempt to use CT scans for the diagnosis of skeletal metastases. Labels are obtained from
\emph{positron emission tomography} imaging (PET). PET/CT registered images are used, where PET high-uptake serves as  ground truth.
We compare our method to Radiomics, common medical imaging features, and to deep neural networks in several configurations.
The main contributions of this work are:
\begin{enumerate}
    \item A general design is proposed for using ensemble classifiers based on spectral total-variation features.
    \item The advantages of this design are shown for the diagnosis of skeletal metastases from CT scans. This is a hard real-world classification problem. The proposed approach significantly outperforms common neural-network methods, as well as classical medical features, indicating that STV representation can be of high relevance.
    \item It is shown that fine scale textures, which can be well visualized by this approach, are of high significance for the classification problem. 
\end{enumerate}



\section{Background}
Positron emission tomography imaging (PET) with 18F-fluoro-2-deoxyglucose (FDG) has become a key method for imaging metabolism in order to detect pathololgical processes (e.g. cancer\ynote{ous} sites) and healthy tissue (e.g. viable heart muscle).  Hybrid PET/x-ray computed tomography (CT) examinations have become the main method in diagnosing skeletal metastases in cancer patients, but these dual imaging modalities expose patients to a larger dose of ionizing radiation than the CT scan itself. The aim of this study is to determine if machine learning methods can be used to extract features from the CT component of the PET/CT imaging that are useful for the diagnosis of skeletal metastases.

Positron emission tomography combined with computed tomography is used in nuclear medicine, particularly to detect cancerous processes. PET/CT imaging is acquired in the same scanner, during a single imaging session \cite{Beyer}, and provides an accurately aligned scan of a high-resolution anatomic image from the CT and a functional image from the PET. The PET modality uses radioactive material that can be absorbed in cancer cells since they are in most of the cases more metabolically active in comparison to the surrounding healthy cells. PET imaging maps body metabolism and allows the detection of disease in its early stages, often before anatomical changes are evident. CT imaging by itself provides a high-resolution anatomic scan but does not give the metabolic information readily provided by PET \cite{townsend}. 

Machine learning (ML) \ynote{has been} widely used in medical imaging over the past two decades. Examples of medical imaging tasks that are solved using machine learning are image segmentation, object detection, CAD diagnosis, classification, image registration, image reconstruction, image denoising and image synthesis. 
Deep learning approaches are commonly used in the medical image segmentation field\ynote{. See for example,} \cite{Zhou}, \cite{ Milletari }, \cite{ Hatamizadeh }, \cite{ Gu }, \cite{ Jha }.  

``Radiomics'' enables high throughput features extraction from CT, MRI or PET images. The extracted data is analyzed for improving oncology patient diagnostic and clinical decision making \cite{Gillies}. There is active research on radiomics  applied in the field of PET-CT imaging. Parmar et al. \cite{Parmar} compared ML approaches and methods of radiomics features selection from pre-treatment CT images of lung cancer patients for achieving accurate and reliable survival prediction. Hsu et al. \cite{Hsu} introduces classifiers based on random forests and FDG-PET radiomics to distinguish normal tissue uptake from uptake in tumor. Bogowicz et al. \cite{Bogowicz} combined CT radiomics features of primary tumors and radiomics of lymph nodes detected by FDG-PET in order to improve loco-regional control prediction accuracy. Bibault et al. \cite{Bibault} used a DNN and pre-treatment radiomics features extracted from CT scan to predict the response after chemoradiation in locally advanced rectal cancer. In recent studies, combination of structural and functional radiomics was explored \cite{Wu}, for example PET/CT radiomics in \cite{Shao}, \cite{Zhou-Wang}, \cite{Pavic}, \cite{Li}, \cite{Xie}.

Artificial intelligence has been applied to the skeletal metastases detection, segmentation, classification, and prognosis. 
Kairemo et al. \cite{Kairemo} conducted a comparative study of PET/CT radiomics analysis for the assessment of the extent of active metastatic skeletal disease in prostate cancer. They concluded that CT alone cannot be used in this task. Yin et al. \cite{Yin} developed MRI-based radiomics model for differentiation of three most common sacrum tumors. Naseri et al. \cite{Naseri}  used planning-CT images of patients with non-metastatic lung cancer and patients with spinal bone metastases to evaluate various feature selection methods and machine learning classifiers for distinguishing between metastatic and healthy bone. Lang et al. presented a convolutional neural network (CNN) that differentiates metastatic lesions in the spine originated from primary lung cancer and other cancers using MRI images and radiomics analysis \cite{Lang}.
Filograna et al. showed the radiomics features extracted from MRI images of 8 oncological patients with bone marrow metastatic disease that is able to differentiate between metastatic and non-metastatic vertebral bodies \cite{Filograna}. Hinzpeter et al. studied the feasibility of differentiation of bone metastases not visible in CT from unaffected bone using [68Ga] Ga-PSMA PET imaging of 67 patients with prostate cancer as ground truth; They selected 11 radiomic features from CT images of 67 prostate cancer patients and trained a gradient-boosted tree to classify patients’ bones into bone metastasis and normal bone, achieving accuracy of 0.85 \cite{Hinzpeter}. Fan et al. \cite{Fan} build classification models utilizing PET texture features to distinguished between benign vs. malignant bone lesions and showed that some texture features showed higher diagnostic value for spinal metastases than SUVmax.

The work of Ben-Cohen et al. \cite{Ben-Cohen} explored PET image synthesis using CT scans and deep learning networks. Their generated virtual PET images can be used for false-positive reduction in lesion detection solutions and may enable lesion detection in a CT-only environment. The lesion detection results were not demonstrated in that work.

Bi et al. \cite{kim} proposed a multi-channel generative adversarial network (M-GAN) for generating PET data. Their model takes the input from the given label and from the CT images, to synthesize the high uptake consistent with the anatomical information. They have shown that their generated PET images boost the training data for machine learning methods. To generate a synthetic PET image using Bi et al. \cite{kim} method, manual labeling of the tumors is required.

\section{Spectral TV}\label{sec:Preliminaries}
In this \ynote{work we} examine the use of \ynote{STV} to provide indicative features for the complex task of uptake prediction based on CT data. STV is an edge-preserving nonlinear transform, \ynote{introduced} in
\cite{gilboa_ssvm,gilboa_siam} and employed for various image-processing tasks, such as image manipulation \cite{Benning}, segmentation \cite{Zeune}, mesh-processing \cite{Fumero} and more
\cite{gilboa_springer}. 
In the context of medical imaging, it has been instrumental \ynote{in} image fusion \cite{hait,Dinh}. The spectral signatures are used here for the first time as classification features. We summarize \ynote{first the fundamental} concept and properties of STV.

\subsection{Spectral Total Variation}\label{subsec:Spectral Total Variation}
Many studies \newtext{have} explored the \newtext{application} of linear models \newtext{in} image processing tasks. The Fourier transform has been the "workhorse" in signal filtering and  representation since it is a mathematically simple, effective, and intuitive method. For real-world piecewise signal with edges this representation is less adequate. Therefore, modern algorithms tend to use nonlinear methods.
Gilboa et al. have conducted research on nonlinear eigenvalue analysis and have introduced a new approach of nonlinear multi-scale representation for image processing \cite{gilboa_springer,gilboa_ssvm,gilboa_siam}. Let us first set \ynote{up} some notations, in order to explain this transform.
The total-variation (TV) energy for smooth functions is
\begin{align}
    \label{eq:TV}
J_{TV}(u) = \int_{\Omega} |\nabla u(x)|dx.
\end{align}
The subgradient $p(u) \in \partial J_{TV}(u)$, where $\partial J_{TV}$ denotes the subdifferential, is
\begin{align}
    \label{eq:p}
 p(u) = -\textrm{div} \left( \frac{\nabla u}{|\nabla u|} \right).
 \end{align}
This definition applies for smooth functions $u$ with non-vanishing gradients and is also referred to as the (negative) 1-Laplacian. A more general definition for signals in the space of bounded variations $BV$ can be found e.g. in \cite{Bellettini,gilboa_springer}.
The transform seeks a decomposition of a signal into elements which admit the following nonlinear eigenvalue problem,
$$ \lambda u \in \partial J_{TV}(u).$$
It turns out that indicator functions of convex sets with low enough curvature (disk-like shapes) admit this eigenvalue problem \cite{Bellettini}, see Fig. \ref{fig:tv_efs}.
The decomposition of a signal $f(x) \in BV$ \newtext{into spectral components} is performed by evolving the total-variation gradient flow 
\begin{align}
    \label{eq:gradientFlow}
    \partial_t u(t,x) \in - \partial J(u(t,x)), \qquad u(0,x)=f(x),
\end{align}
where $\partial_t$ denotes the first partial derivative with respect to an artificial time parameter $t$. This parameter can be also interpreted as scale in nonlinear scale-space.
The spectral component (i.e. the transform) of $f(x)$ at scale $t$ is defined by
\begin{align}
\label{eq:phi}
 \phi(t,x) := t\partial_{tt}u(t,x),
\end{align}
where $u(t,x)$ is the solution of \eqref{eq:gradientFlow} and $\partial_{tt}$ is the second order partial derivative w.r.t. $t$. 
In \cite{burger} the following key properties were shown:
\begin{enumerate}
  \item A single eigenfunction with eigenvalue $\lambda$ appears as a delta at scale $t=1/\lambda$ in the transform domain.
  \item The reconstruction formula for decomposed data $f$ is based on simple integration of the spectral components via
  \begin{align}
    \label{eq:reconst}
    f =  \int_0^\infty \phi(t) ~dt + \bar{f},
\end{align}
   where $\bar{f}$ is the mean value of $f$. 
  \item A spectrum is defined by 
  \begin{align}
    \label{eq:S}
    S(t) := \langle f,\phi(t)\rangle_\Omega,
\end{align} 
  which admits a Parseval-type identity, $\int_0^\infty S(t)dt = \|f\|^2_{L^2}$.
  \item Motivated by the reconstruction formula \eqref{eq:reconst}, filtering is formulated in an analog\ynote{ous} manner to Fourier filtering via
\begin{align}
\label{eq:filter}
f_H =  \int_0^\infty H(t) \; \phi(t) ~dt + \bar{f},
\end{align}
for a filter function $H$ that can enhance, damp or remove certain frequencies.
\end{enumerate}
We note that the above can be generalized beyond TV, for all norms and semi-norms in a discrete setting within Euclidean spaces or graphs. In a restricted setting, such as the discrete 1D case, it was shown in 
\cite{burger} that:
\begin{enumerate}
  \item The spectral representations yield a decomposition of the input signal into eigenfunctions.
  \item Spectral components (which turns out to be a difference of two eigenfunctions) are pairwise orthogonal.
\end{enumerate}
Additional details o\ynote{f} the transform and its \ynote{implementations} can be found in \cite{gilboa_siam,gilboa_springer}.

\textbf{Computational considerations.} We note that computing STV involves solving a nonlinear PDE (TV-flow) which requires some computational efforts. \newtext{At each scale, the solution to the TV flow must be computed, involving solving multiple non-smooth optimization problems. This process can be computationally expensive}.  There are several studies which aim at considerably increasing the speed of these computations (by orders of magnitude), most notably by specific neural network architectures, see \cite{grossmann2020deeply,grossmann2022unsupervised,langer2024deeptv}.

\subsection{Spectral TV Local Scale Signatures}\label{subsec:Spectral TV local scale signatures}
Hait et al. \cite{hait} used STV to form a continuous, multi-scale, fully edge-preserving, and local descriptor, referred to as spectral total-variation local scale signatures. 
Given an image $f(x)$, we denote by $\phi_f(x,t)$ the respective STV spectral components, Eq. \eqref{eq:phi}.
A \ynote{time-discrete} setting of the TV gradient flow, \eqref{eq:gradientFlow}, is solved. The evolution is for $n$ iterations with time step $\Delta t$, reaching a maximal scale of $T=n \cdot \Delta t$. 
We thus obtain for each pixel a vector of $n$ features, as shown in Fig. \ref{fig:feat_ext} top right.
This feature vector (signature) can be viewed as the multiscale spectral response of the pixel.
Several properties of these signatures were presented in  \cite{hait}:

\emph{Sensitivity to size and to local contrast.}
Objects with different contrast and size have distinct signatures \ynote{in their image}. Spatial scaling and contrast change by a positive scalar $a$ affect the signatures \ynote{as follows},
\begin{equation} \label{eq6}
\phi_{f(ax)}=a\phi_f (ax,at) ,
\end{equation}
\begin{equation} \label{eq7}
\phi_{af(x)}=\phi_f (x,t/a) .
\end{equation}


\emph{Invariance to Rotation, Translation, and Flip.}
The following invariances in $\mathbb{R}^n$ were shown,
\begin{equation} \label{eq9}
\phi_{f(Rx)}=\phi_f (Rx,t),
\end{equation}
\begin{equation} \label{eq10}
\phi_{f(x-a)}=\phi_f (x-a,t),	
\end{equation}
\begin{equation} \label{eq11}
\phi_{f(x)}  (x)=\phi_{f(-x)}(-x).	
\end{equation}
where \newtext{$R$} is a rotation matrix, and $a$ is a spatial shift.
Hait et al. presented several applications \ynote{of} the proposed framework, such as image clustering and fusion of thermal, optic, and medical images. 
To increase the distinctness of salient objects  and to improve clustering, the following signature enhancement was \ynote{proposed},
\begin{equation} \label{eq12}
\Phi_f (x)=\phi(x,t)\cdot\|\phi(x,t)\|_{[0,T]}^p.
\end{equation}
\newtext{This enhancement follows the methodology introduced in \cite{hait}, where each signature is “stretched” according to its $L_p$ norm (commonly $L_1$), which strengthens and sparsifies the representation of salient structures. This process increases the separation between different features, thus improving distinctness and clustering performance.}

\section{Classification Method}\label{sec:Methods}
We investigate a new method of diagnosing patients with skeletal metastases, based on \ynote{} CT imaging and STV, without PET metabolic function imaging. 
An STV transform of CT images allows extracting local descriptors that potentially uncover conspicuous pathology that may not be noticeable in the CT scan. The unique features at the locations of pathology were confirmed by the diagnostic readings of the PET radiotracer uptake. 

The proposed method should enable the detection of pathology based on CT scans consistent with metabolic scans\ynote{. It could as such} aid in patient diagnosis and management. Our framework includes training and test modules. The training is performed using CT images and lesion labels acquired through PET. The test process is based on CT images only, which may allow diagnosis in early stages of pathology.

 We compare STV to radiomics features, both methods are followed by classical learning (based on decision trees). In addition, this study is complemented by examining deep neural networks \ynote{which execute} the same task. Since data is limited, to train neural networks in an optimal manner, transfer learning is used, as customary in such cases. Two pretrained networks were examined, based on natural images and on medical images. The training and testing sets for all methods were the same.






\subsection{PET/CT imaging and image acquisition}
\label{subsec:Image Acquisition}

Patient scans were performed on a PET/CT system (Discovery 690, GE Healthcare, Waukesha, WI) combining a dedicated time-of-flight PET with a lutetium-yttrium oxyorthosilicate (LYSO) scintillator and a 64-slice CT. The institutional review board approved this retrospective study, waiving the requirement to obtain informed consent. After an intravenous injection of an activity of 5 MBq/kg FDG, a 60 minute uptake period was followed by a PET acquisition of 1.5 minutes per bed position. Patient scans were reconstructed on a 256 × 256 matrix, with a 70 cm transverse field-of-view (FOV) and a 3.27 mm slice thickness using a 3D ordered subset expectation maximization (OSEM) algorithm (24 iterations, 2 subsets, 6.4 mm full width at half maximum Gaussian filter). The CT was acquired using a peak voltage of 120 kV, a pitch of 1.375, 0.7 s/rotation, and variable current (100 mA minimum, 650 mA maximum, noise index 11.57). CT images were reconstructed on a 512 × 512 matrix, with a 50 cm diameter transverse FOV and a 2.5 mm slice thickness using filtered backprojection and 40\% adaptive statistical iterative reconstruction.

\subsection{Construction of experimental data set}
We analyzed 457 patients: 249 oncological patients with confirmed bone metastases disease and 208 patients without metastases. Our focus was to examine the lower spine column. 
Among 249 oncological patients we identified  71 patients that
were diagnosed with vertebral lesions in the lower spine column and included them in our study.  Other oncological patients were identified with bone metastases not in the spine or not in the lower spine column. We then selected 71 healthy patients to be included  in our study as a way of keeping a balanced number of pathological and healthy patients. In total,  our model was built on 142 patients with 71 lesions in the lower spine and 71 without lesions. 

From 71 pathological patients we extracted 613 patches of lower spine with high uptake regions and 298 patches of lower spine with low uptake regions. From 71 healthy patients we extracted 613 patches of the lower spine. To reduce the imbalance we used the low uptake patches of pathological patients twice, so in total we used 1822 (1524 unique) two-dimensional patches for building our framework (613 patches with high uptake, 596 patches from pathological patients with low uptake, 613 patches of healthy patients).

\subsection{Data pre-processing}
\label{subsec:Data pre-Processing}
\begin{figure}[htb]
\centering 
\includegraphics[width=3.49in]{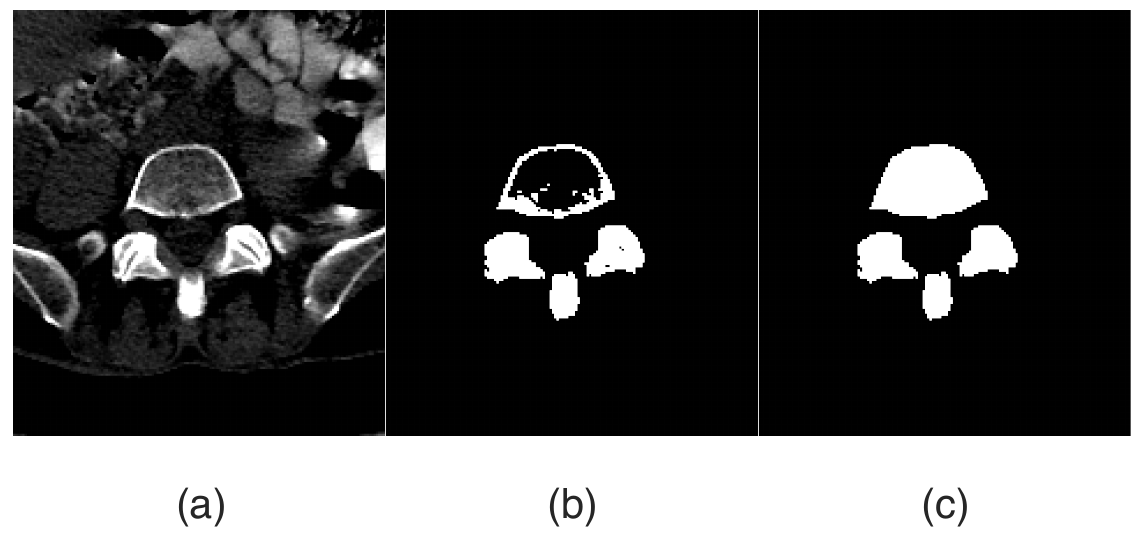}
\caption{Segmentation of a vertebra patch. (a) CT image. (b) Slice from a 3D binary mask obtained with a sparse field method \cite{Whitaker}. (c) The holes in the binary mask (b) are filled using a flood-fill algorithm.}
\label{fig:segmentation}
\end{figure}

Our data set includes a set of PET/CT 3D volumes of patients diagnosed with skeletal metastases and PET/CT 3D volumes of healthy patients.  
In order to obtain a 3D spine mask of each 3D CT volume, we utilized the spine segmentation methodology of Khandelwal et al. \cite{pulkit}. Using the obtained 3D spine mask and morphological operations, we extracted CT patches of vertebrae from the lower spine column, obtaining CT patches of 5 lumbar vertebrae (L1-L5) and last thoracic vertebra (T12). From each vertebra we extract up to 6 patches. An example of vertebra segmentation is shown in Fig. \ref{fig:segmentation}. The extracted patches' size was 50 × 50 pixels and a center of a patch was from the center of a vertebra's body.  
The obtained patches were manually examined, and if a patch did not include the vertebra due to incorrect segmentation, it 
was removed from the data set. For each patch in the obtained data set, we extracted the registered PET patch from the corresponding PET volume.

\subsection{Data labeling}\label{subsec:Data Labeling}
Each patient in the data set was labeled by a certified specialist nuclear medicine physician, providing a data-set with studies labeled as “Normal” and “Pathological”.
All patches that were extracted from the normal studies were labeled as “Normal”.
We used a set of PET patches from our pre-processed data set to label CT patches that were extracted from the pathological studies. 
We applied a circular mask of 13 pixels around the center of each PET/CT patch. In \ynote{cases} where the PET patch displayed a high uptake within the circular mask, we labeled the CT patch as “Pathological High Uptake (HU)”, otherwise we labeled \ynote{them} as “Pathological Low Uptake (LU)”. \newtext{From 71 pathological patients, we extracted 613 patches labeled as ``Pathological High Uptake (HU)'' and 298 patches labeled as ``Pathological Low Uptake (LU).'' From 71 healthy patients, we extracted 613 patches labeled as ``Normal.'' }

\subsection{Proposed framework}\label{subsec:Proposed framework}
\begin{figure}[ht]
\centering 
\includegraphics[width=3.49in]{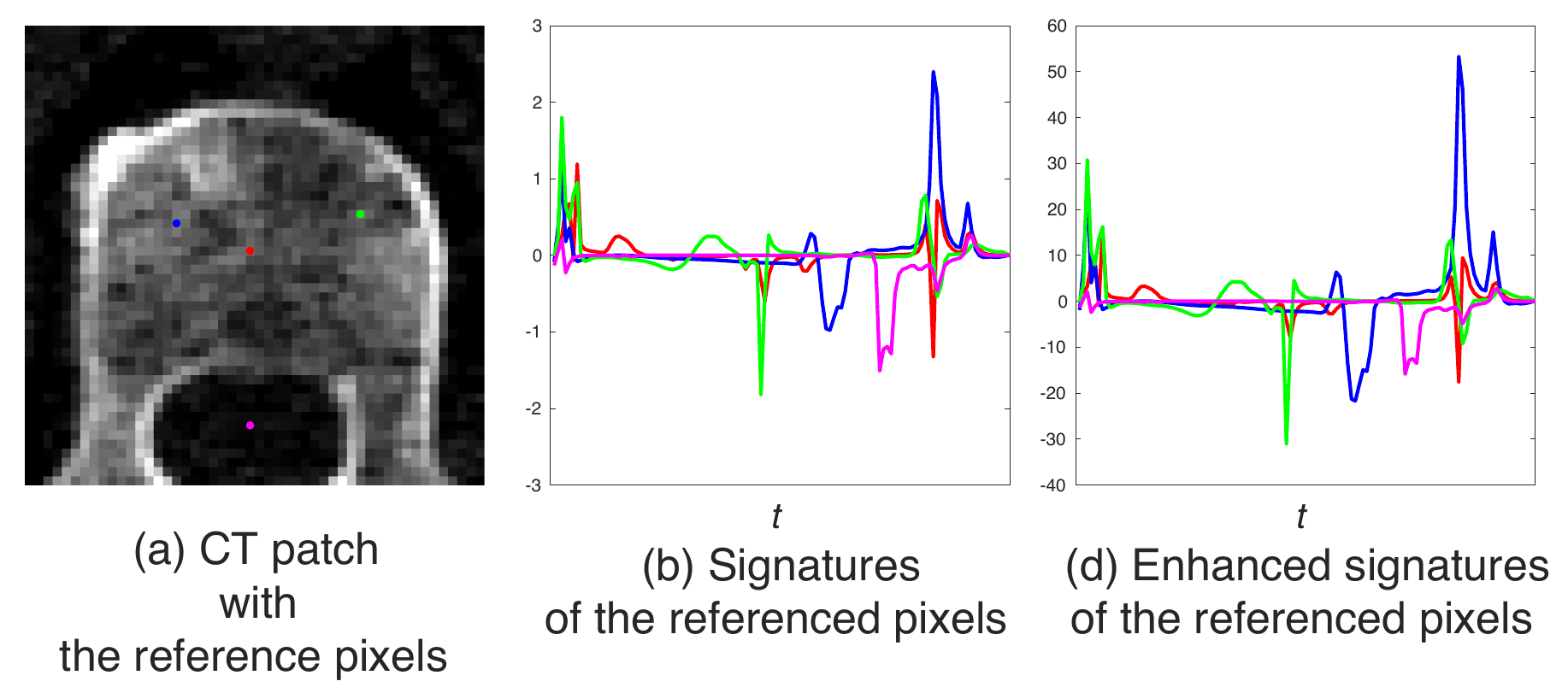}
\caption{Distinctness of signatures of the referenced pixels.}
\label{fig:enhanced_sig}
\end{figure}

\begin{figure}[ht]
\centering 
\includegraphics[width=3.49in]{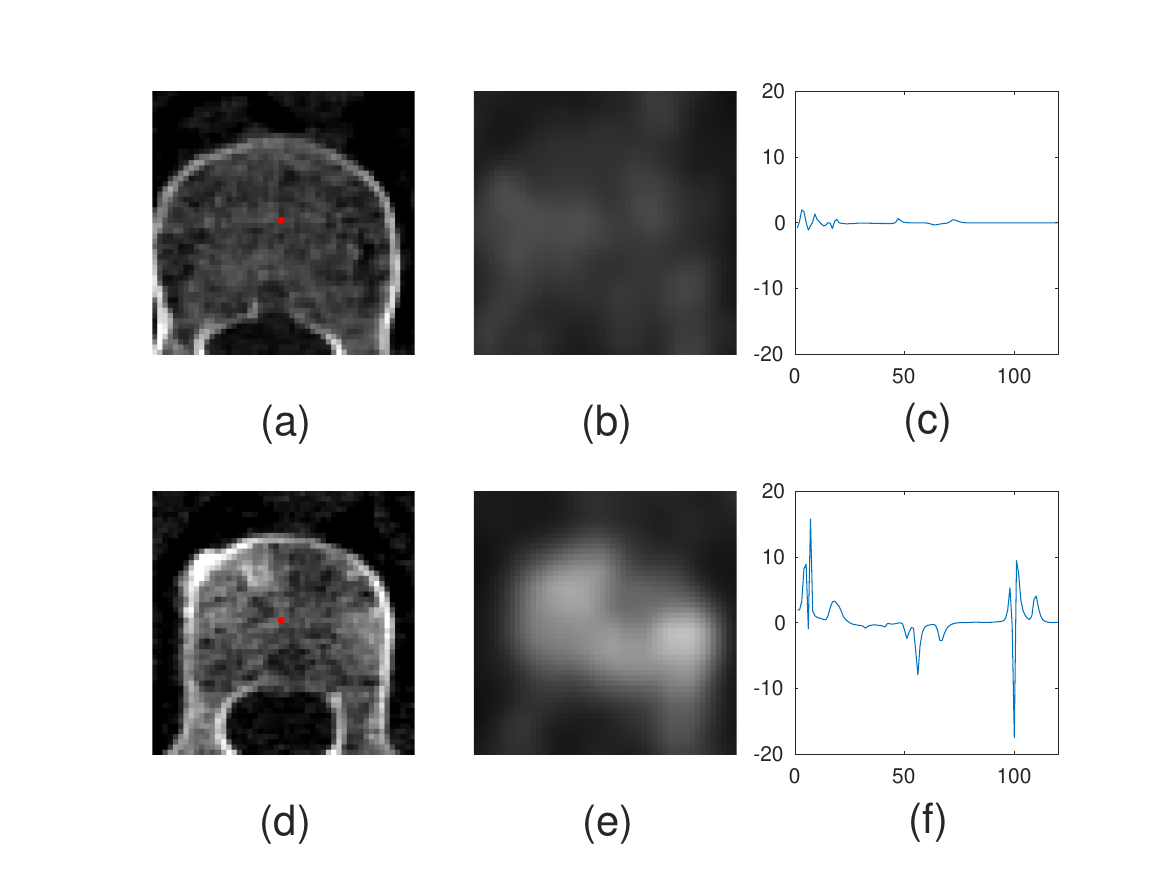}
\caption{Example of CT and PET patches and signature of the referenced pixel (red). (a) Normal CT patch with referenced pixel. (b) Normal PET patch. (c) Normal spectral TV local scale signature of referenced pixel in CT patch (red). (d) Pathological CT patch with reference pixel. (e) Pathological PET patch. (f) Pathological spectral TV local scale signature of referenced pixel in CT patch.   }
\label{fig:ct_pet_sig_example}
\end{figure}

\begin{figure*}[htb]
\centering 
\includegraphics[width=5.14in]{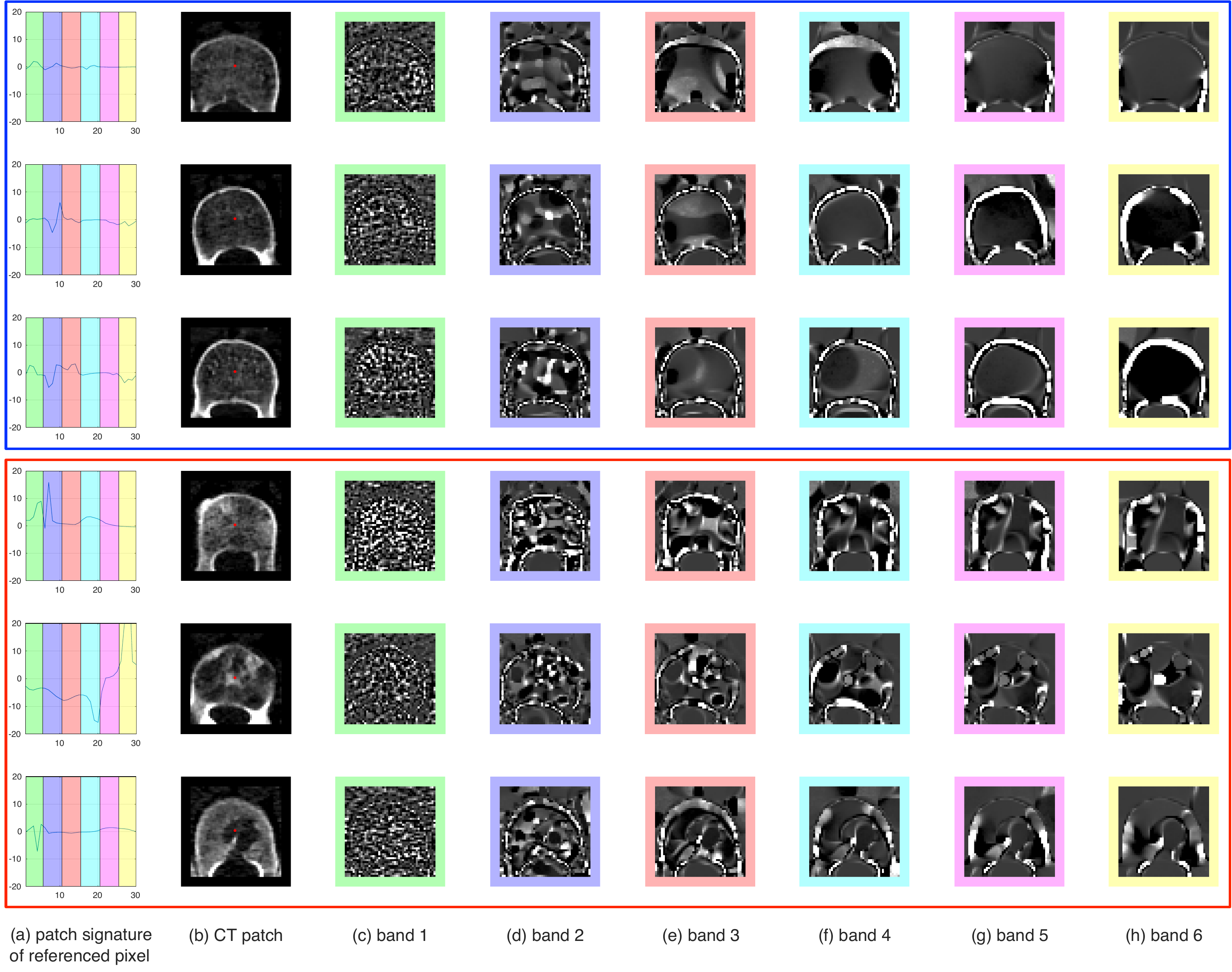}
\caption{Visual demonstration and comparison of normal and pathological bands. Normal examples are shown in the blue frame. Pathological examples are shown in the red frame.}
\label{fig:vis_bands3}
\end{figure*}

After applying the spectral TV transform on a CT patch, using a signature enhancement with $p=1$ (Eq. \eqref{eq12}), we obtain 120 STV components of a patch. \newtext{We selected p = 1 for signature enhancement following the methodology proposed in \cite{hait}, where it was shown to improve distinctness and sparsity of salient structures.}
Examples of distinctness of different spectral signatures (non-enhanced and enhanced) coming from one CT patch are presented in Fig. \ref{fig:enhanced_sig}. 
Enhanced spectral signatures extracted from normal and pathological CT patches are presented in Fig. \ref{fig:ct_pet_sig_example}. 
We \ynote{then apply} a circular mask of 13 pixels around the center of each component, and use the obtained circles as input to the proposed framework. 
Every 5 adjacent components \ynote{are} considered a band, producing 24 bands in total, see Fig. \ref{fig:feat_ext}. 
We rely on the fact that the spectral bands are lowly correlated. Theoretically, in 1D we obtain precise orthogonality and in 2D and higher dimensions this orthogonality is still an open problem, however numerically we observe lowly correlated bands \cite{gilboa_springer}. For visualization purposes we present the normal and pathological first 6 bands in Fig. \ref{fig:vis_bands3}.

\begin{figure}[ht]
\centering
\subfloat{%
       \includegraphics[width=3.49in]{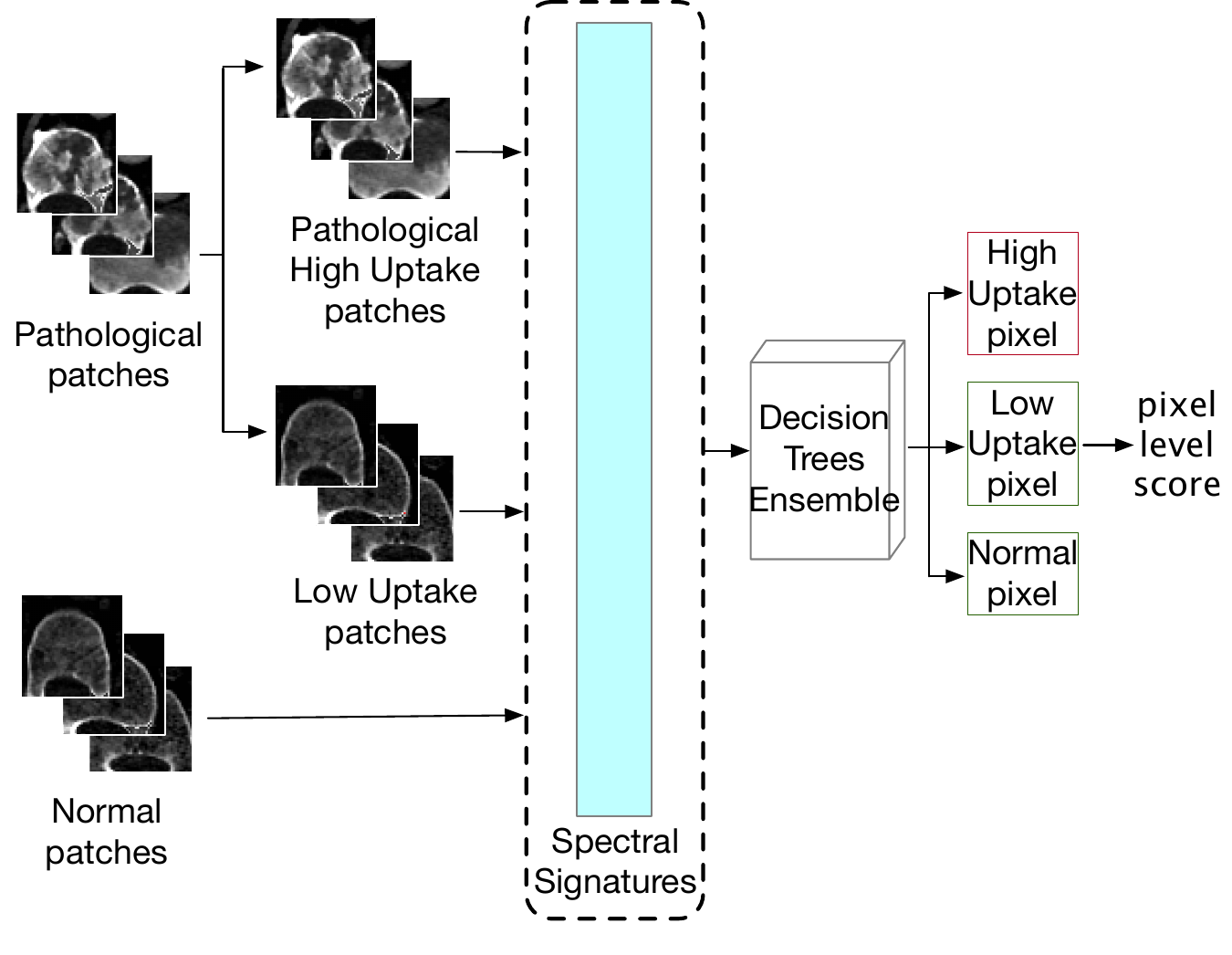}}
\caption{Overview of the training procedure for pixel level prediction. First Spectral signatures were obtained by applying STV transform on the input CT patches. The spectral signatures of the pixels inside the circle mask were deployed as features for training Decision Tree Ensemble, which generates 3-class labels. The "Low Uptake" label was replaced with the "Normal" label, and the pixel level score was calculated.}
\label{fig:train}
\end{figure}

\begin{figure}[ht]
\centering
\subfloat{%
       \includegraphics[width=3.49in]{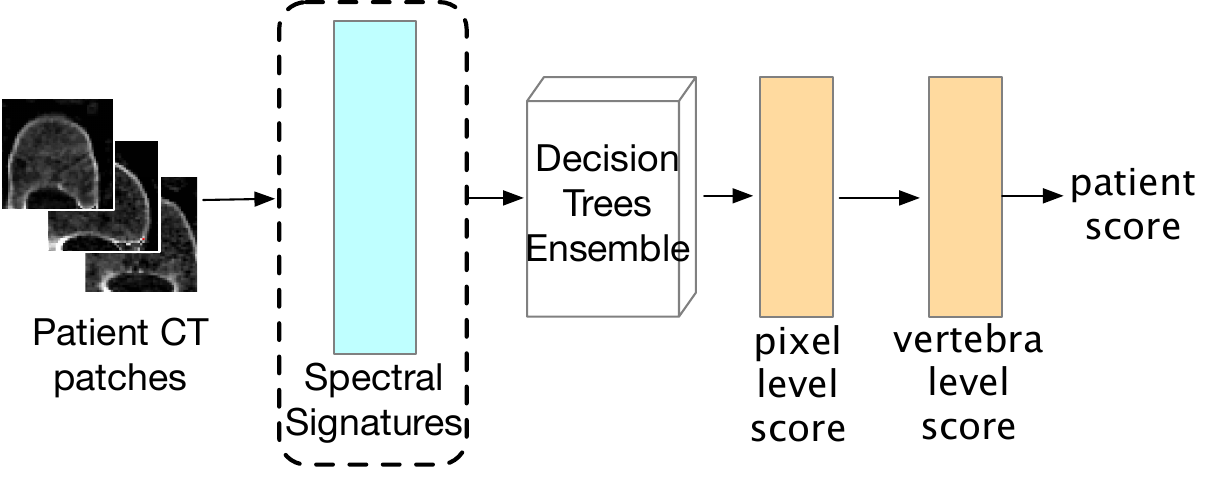}}
\caption{Overview of the testing procedure for patient level prediction. First Spectral signatures were obtained by applying STV transform on the input CT patches. The spectral signatures of the pixels inside the circle mask were deployed as features for testing Decision Tree Ensemble, which generates 3-class labels. The "Low Uptake" label was replaced with the "Normal" label, and the pixel level score was calculated. The vertebra score was calculated by averaging pixels' scores that belong to one vertebra. The patient is labeled as "Pathological" if he has at least one vertebra with a score higher than some predefined threshold.}
\label{fig:test}
\end{figure}

The model \ynote{is} composed of 24 
decision trees, where each tree is trained using its associated band of features. For example, the decision tree number 1 was trained using the feature band number 1, which included scales from 1 to 5, the decision tree number 2 was trained using the feature band number 2, which included scales from 6 to 10, etc. Each decision tree was trained \newtext{using MATLAB’s \texttt{fitctree} function with default hyperparameters, and was tasked with classifying} the signature’s band into 3 classes i.e., “Pathological High Uptake”, “Pathological Low Uptake”, and “Normal”. Each pixel was characterized by 24 bands, that is, the framework generated  24 tags for each pixel. Our purpose is to reveal high uptake regions. Thus, after each pixel was classified with 24 tags, the “Pathological Low Uptake” tags were \newtext{changed} with the tag “Normal” to \ynote{convert} the problem into a binary classification problem. The prediction score of the pixel was obtained by calculating the mean of 24 tags that \ynote{were} associated with the pixel. Averaging the pixels’ scores of a patch, provided a patch score. Each vertebra consisted of at most 6 patches, that is, the vertebrae’s score was obtained by averaging its patches scores. If at least one vertebrae score was higher than a predefined cutoff the study was classified as “Pathological”. 
Our train and test flowcharts for decision trees based model are presented in Fig. \ref{fig:train} and Fig. \ref{fig:test}.

Trying to enhance the \ynote{quality of the} results we also utilized the random forest algorithm per each band of features. We generated 24 random forests \ynote{and} each forest was trained on a different band of features.
In the results section we can observe that the use of random forest, in this case, is not essential.

\begin{algorithm}[H]
	\caption{Algorithm for Patient Diagnosis}
	\begin{algorithmic} [1] 
		\For {Every $fold$} 
            \State $Xtest\leftarrow fold_i$ \Comment{Keep the current $fold$ as a hold-out} 
            \State $Xtrain\leftarrow$ $remaining$ $folds$ \Comment{Use the remaining folds as a training data-set} 
            \For {Every $feature$ $band$}
                \State $Xtrain\_band\leftarrow Xtest(feature$ $band_j)$
				\State $fit(Xtrain\_band)$ \Comment{Fit the model on the features band from the training data set}
                \State $Xtest\_band\leftarrow Xtest(feature$ $band_j)$
				\State $Ytest\_predicted\_band=predict(Xtest\_band)$ 
    
    \Comment{Test the model on the features band from the hold out fold}
                \State $Ytest\_predicted=[Ytest\_predicted$ $Ytest\_predicted\_band]$ \Comment{Retain the prediction results}
			\EndFor
		\EndFor
        \For {Every $patient$}
                \For {Every $low\ back\ vertebra$}
                \State Calculate $ vertebra\ mean\ prediction\ P_{v_{i}}$
                \If{ $P_{v_{i}} > cutoff$}
                \State Label patient as "Pathological"
                \EndIf
                \EndFor
                \EndFor      
	\end{algorithmic} 
\end{algorithm}

Our algorithm can be interpreted as an ensemble method, where for each patch we have an ensemble of size 24. Since the bands are lowly correlated, the ensemble members produce diverse predictions, yielding a very powerful classifier that can accommodate well noisy and unreliable data. As we see later in the experimental section, this is a hard classification problem and the theoretical knowledge on STV affords a considerable improvement over generic methods.

\subsection{Ablation study}\label{subsec:Ablation}

In the ablation study, we changed the number of adjacent components (scales) that were united into each band\ynote{. We thereby} changed the total number of bands. The ablation study was performed on more restricted dataset that included 57 healthy patients without lesions and 57 pathological patients with confirmed lesions in the lower spine column. Different values of the adjacent components per band and the matching results are shown in Table \ref{tab:abl_table}.


\makeatletter
\newcommand{\thickhline}{%
    \noalign {\ifnum 0=`}\fi \hrule height 1pt
    \futurelet \reserved@a \@xhline
}
\newcolumntype{"}{@{\hskip\tabcolsep\vrule width 1pt\hskip\tabcolsep}}
\makeatother

\begin{table*}[htbp]
\caption{Performance Comparison, 10 Fold Cross Validation} 
\label{compare_table2}
\centering
\resizebox{\textwidth}{!}{ 
  \begin{threeparttable}
    \begin{tabular}{|cV{3.5}cV{3.5}c|c|c|c|c|c|}
    \hline
    Classification &  &
    \multicolumn{4}{c|}{Other Metrics}\\ 
    \cline{3-6}
    Method   & AUC & Accuracy & Specificity & Recall & Precision   \\ 
    \hline
    Decision Trees Ensemble (ours) & \textbf{0.87\textpm 0.01} & \textbf{0.80\textpm 0.01} & \textbf{0.84\textpm 0.04} & \textbf{0.77\textpm 0.04} & \textbf{0.83\textpm 0.03}  \\ 
    \hline
    Random Forests Ensemble (ours) & \textbf{0.87\textpm 0.01}   & \textbf{0.80\textpm 0.01} & {0.83\textpm 0.04} & \textbf{0.77\textpm 0.05} & {0.82\textpm 0.02}  \\ 
    \hline
    Random Forest, Spectral TV features & 0.67\textpm 0.03  & 0.65\textpm 0.02 & 0.69\textpm 0.05 & 0.60\textpm 0.05 & 0.66\textpm 0.03  \\
    Random Forest, Radiomics Features & 0.79\textpm 0.01  & 0.73\textpm 0.01 & 0.71\textpm 0.05 & 0.75\textpm 0.05 & 0.73\textpm 0.02 \\
    Random Forest Ensemble, Radiomics Features & 0.79\textpm 0.004  & 0.74\textpm 0.01 & 0.68\textpm 0.01 & 0.81\textpm 0.02 & 0.71\textpm 0.01 \\
    \hline
    {Transfer Learning ResNet-50, ImageNet \cite{Deng}, freezeall} & 0.60\textpm 0.01  & 0.62\textpm 0.02 & 0.45\textpm 0.07 & 0.78\textpm 0.07 & 0.59\textpm 0.01  \\
    {Transfer Learning ResNet-50, ImageNet \cite{Deng}, unfreezetop10} & 0.62\textpm 0.05  & 0.62\textpm 0.04 & 0.64\textpm 0.12 & 0.61\textpm 0.12 & 0.63\textpm 0.04  \\
    {Transfer Learning ResNet-50, ImageNet \cite{Deng}, additional layers} & 0.70\textpm 0.02  & 0.68\textpm 0.02 & 0.61\textpm 0.06 & 0.75\textpm 0.04 & 0.66\textpm 0.03 \\
    \hline
    {Transfer Learning ResNet-50, RadImageNet \cite{Mei}, freezeall} & 0.75\textpm 0.01  & 0.69\textpm 0.01 & 0.75\textpm 0.09 & 0.64\textpm 0.08 & 0.73\textpm 0.05  \\
    {Transfer Learning ResNet-50, RadImageNet \cite{Mei}, unfreezetop10} & 0.68\textpm 0.02 & 0.65\textpm 0.02 & 0.82\textpm 0.17 & 0.48\textpm 0.18 & 0.79\textpm 0.13   \\
    {Transfer Learning ResNet-50, RadImageNet \cite{Mei}, additional layers} & 0.68\textpm 0.02  & 0.66\textpm 0.01 & 0.75\textpm 0.12 & 0.56\textpm 0.12 & 0.70\textpm 0.06  \\
    \hline
    \end{tabular}  
  \end{threeparttable}
  }
\end{table*}

\begin{table*}[htbp]
\caption{Ablation Study, 10 Fold Cross Validation} 
\label{tab:abl_table}
\centering
\resizebox{\textwidth}{!}{ 
  \begin{threeparttable}
    \begin{tabular}{|c|c|c|c|c|c|c|c|c|c|}
    \hline
    Classification & Scales per & Correlated &
    \multicolumn{6}{c|}{Metric}\\ 
    \cline{4-9}
    Method & tree &  & Cutoff & Accuracy & Specificity & Recall & Precision & AUC  \\ 
    \hline
    Decision Trees Ensemble & 3 & No & 0.45 & 0.78 & 0.84 & 0.72 & 0.82 & 0.84  \\ 
    Decision Trees Ensemble & 4 & No & 0.45 & 0.79 & 0.74 & 0.84 & 0.76 & 0.86  \\ 
    Decision Trees Ensemble & 5 & No & 0.45 & \textbf{0.81} & 0.76 & 0.86 & 0.78 & \textbf{0.87} \\ 
    Decision Trees Ensemble & 6 & No & 0.45 & 0.79 & 0.76 & 0.82 & 0.77 & 0.86 \\ 
    Decision Trees Ensemble & 8 & No & 0.5 & 0.78 & 0.9 & 0.66 & 0.87 & 0.86  \\ 
    Decision Trees Ensemble & 10 & No & 0.5 & 0.76 & 0.86 & 0.66 & 0.82 & 0.84  \\ 
    Decision Trees Ensemble & 12 & No & 0.5 & 0.77 & 0.84 & 0.7 & 0.81 & 0.83  \\ 
    Decision Trees Ensemble & 15 & No & 0.5 & 0.7 & 0.76 & 0.64 & 0.73 & 0.79  \\ 
    \hline
    Decision Trees Ensemble & 3 & Yes & 0.5 & 0.7 & 0.5 & 0.9 & 0.6428 & 0.8 \\ 
    Decision Trees Ensemble & 4 & Yes & 0.5 & 0.69 & 0.48 & 0.9 & 0.63 & 0.79  \\
    Decision Trees Ensemble & 5 & Yes & 0.55 & 0.72 & 0.72 & 0.72 & 0.72 & 0.8  \\
    Decision Trees Ensemble & 6 & Yes & 0.5 & 0.69 & 0.48 & 0.9 & 0.63 & 0.79 \\
    Decision Trees Ensemble & 8 & Yes & 0.5 & 0.69 & 0.48 & 0.9 & 0.63 & 0.79  \\
    \hline
    Random Forests Ensemble & 5 & No & 0.5 & \textbf{0.81} & 0.8 & 0.82 & 0.8 & \textbf{0.87}  \\ 
    Random Forests Ensemble & 6 & No & 0.5 & 0.75 & 0.68 & 0.82 & 0.72 & 0.85  \\ 
    \hline
    Random Forests Ensemble & 3 & Yes & 0.75 & 0.7 & 0.82 & 0.58 & 0.76 & 0.77  \\ 
    Random Forests Ensemble & 5 & Yes & 0.75 & 0.7 & 0.82 & 0.58 & 0.76 & 0.77  \\ 
    \hline
    Random Forest & 120 & N.A. & 0.9 & 0.61 & 0.6 & 0.62 & 0.6 & 0.61  \\
    \hline
    \end{tabular}
  \end{threeparttable}
  }
\end{table*}

\section{Results}\label{sec:Results}
Our proposed model was trained and validated using a 10-fold cross-validation technique. The validation was done on the out-of-fold patients, that is, on patients that were not used to train the model. 
We evaluated the proposed framework for validation via area under curve (AUC) of the receiver operating characteristic (ROC). Table \ref{compare_table2} shows that the AUC of the proposed framework \ynote{yields} over 0.87.
To avoid class imbalance, we selected a similar number of CT patches from different classes.

\begin{figure}[H]
\centering 
\includegraphics[width=3.49in]{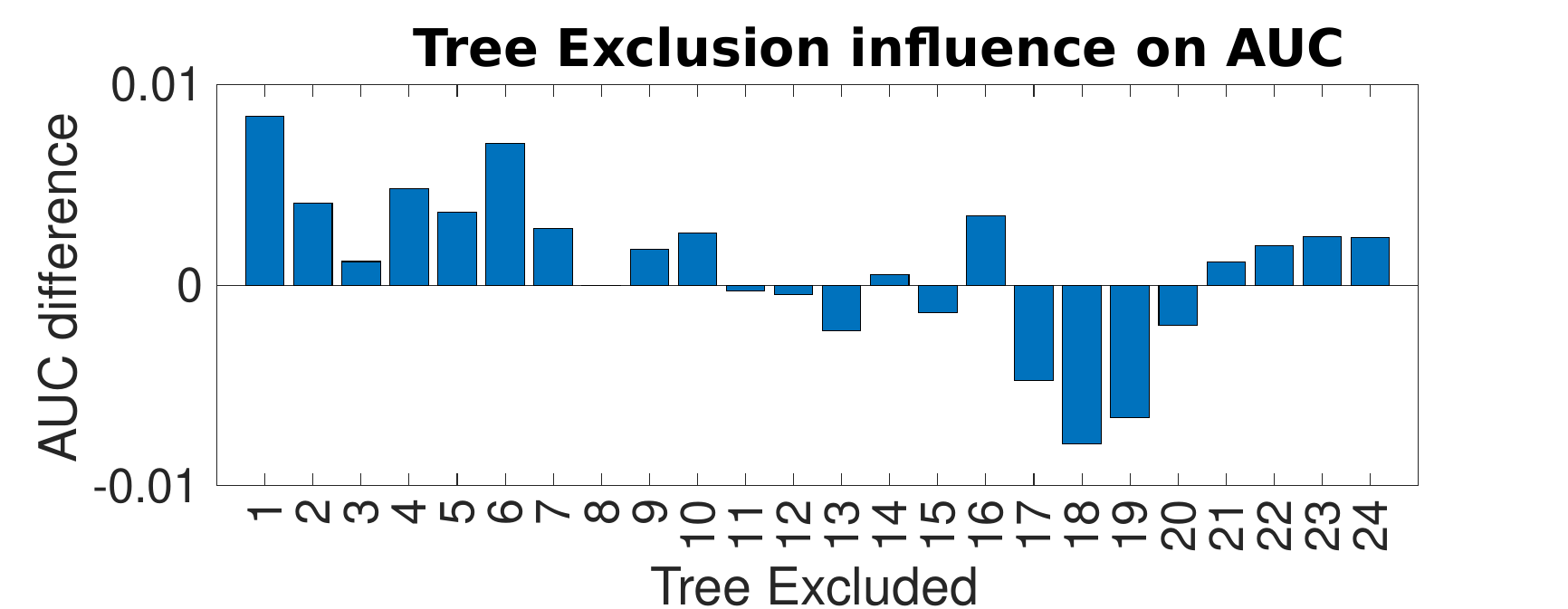}
\caption{
{\bf Band importance.} In this experiment we attempt to assess the importance of each band for the final classification. This is done by excluding the tree corresponding to a band (out of 24 bands) and using the other 23 bands only for classification. We show the AUC drop compared to the AUC of using all bands $AUC_{orig}-AUC_{exc-band}$ (positive means the band is contributing positively). Trees associated with finer scales are generally more important for the final classification results. 
}
\label{fig:auc_inf2}
\end{figure}

\begin{figure}[H]
\centering 
\includegraphics[width=3.49in]{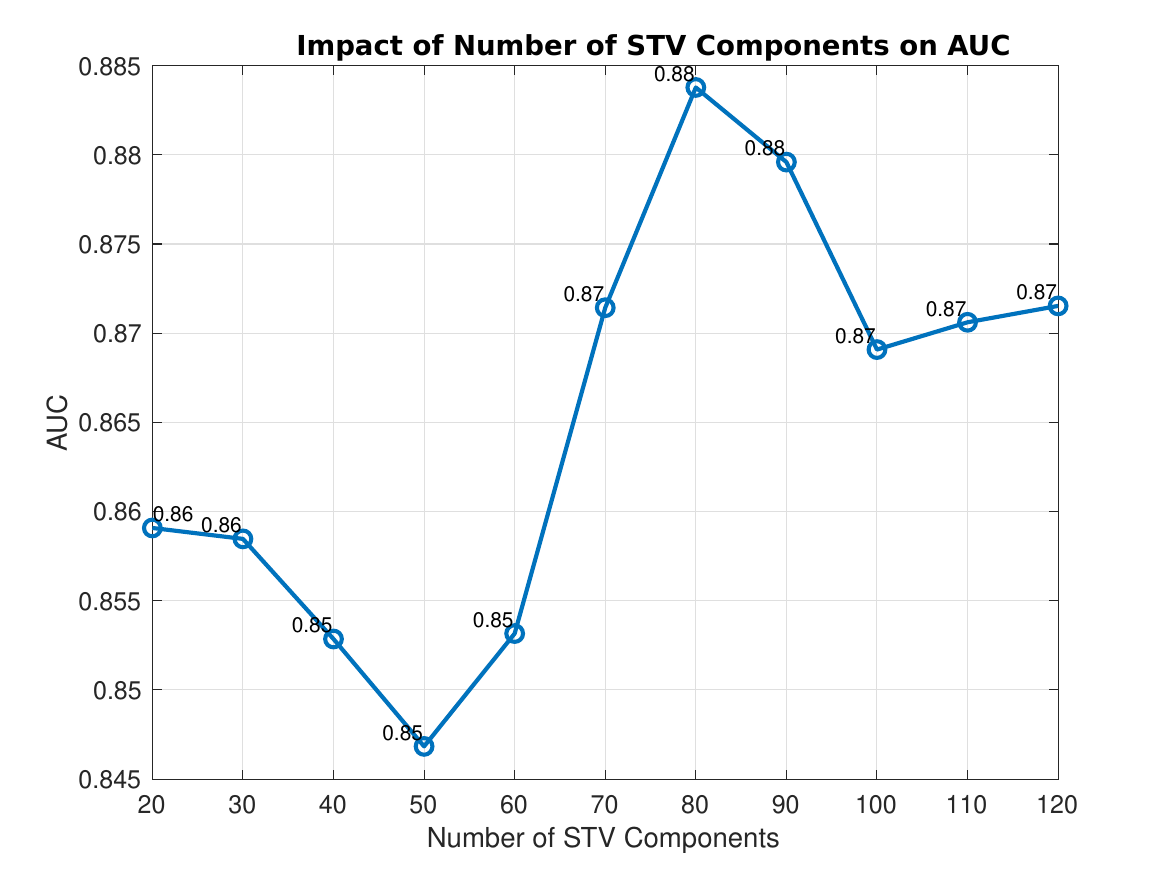}
\caption{
{\bf The impact of the number of STV components.} The classifier's performance, measured by AUC, as a function of the number of STV components used in the classification task.
}
\label{fig:stv_auc}
\end{figure}

To investigate the importance of each decision tree in the ensemble, we exclude one decision tree and evaluated the performance in terms of AUC, based on the other 23 trees. The AUC drop is shown in Fig. \ref{fig:auc_inf2}).  This experiment shows  that trees that are built based on fine scale features are generally contributing more to the framework’s performance. \newtext{Interestingly, we also observed that the exclusion of certain trees led to an increase in AUC, particularly starting from band number 17 (corresponding to STV components from 81 onward).
Additionally, we systematically examined the impact of the number of STV components used in the classification. We varied the number of components from 20 to 120 in steps of 10 and measured the corresponding AUC. The results, presented in Fig. \ref{fig:stv_auc}, show that using 80 STV components yielded the highest AUC (0.88), slightly outperforming the 120-component configuration (0.87). This finding aligns with our tree exclusion analysis, suggesting that some higher-band STV components may introduce noise rather than useful information.}

To explore a generic learning approach for classification, the random forest (RF) model was built based on all 120 STV components. Here we do not use ensembles and do not incorporate our prior knowledge on STV's low correlated nature. The performance thus drops sharply, where the RF algorithm yields the AUC value of 0.67.
\newtext{We did not evaluate boosting-based classifiers in this study. Investigating whether boosting methods could further improve classification performance is a promising direction for future work.}

\newtext{To further motivate the choice of decision trees, we also evaluated a simple multi-layer perceptron (MLP) model directly on the spectral signatures. This model, which included three hidden layers with 15, 10, and 5 neurons respectively, achieved an AUC of only 0.63, substantially lower than our proposed framework.}

Next, we compared our proposed framework to the radiomics features based model. In order to extract radiomics features from the CT patches, we used the open-source PyRadiomics package \cite{Griethuysen}.
From each CT patch, we extracted 93 radiomics features (18 first order, 24 gray level co-occurrence matrix (GLCM), 14 gray level dependence matrix (GLDM), 16 gray level run length matrix (GLRLM), 16 gray level size zone matrix (GLSZM), 5 neighbouring gray tone difference matrix (NGTDM) features), using the same circle mask. Similarly to our method, we used 10 fold cross validation to train and evaluate random forest classifier based on the extracted radiomics features. The yielded AUC value is 0.79, the results are presented in table \ref{compare_table2}.

\begin{figure}[H]
\centering 
\includegraphics[width=3.4in]{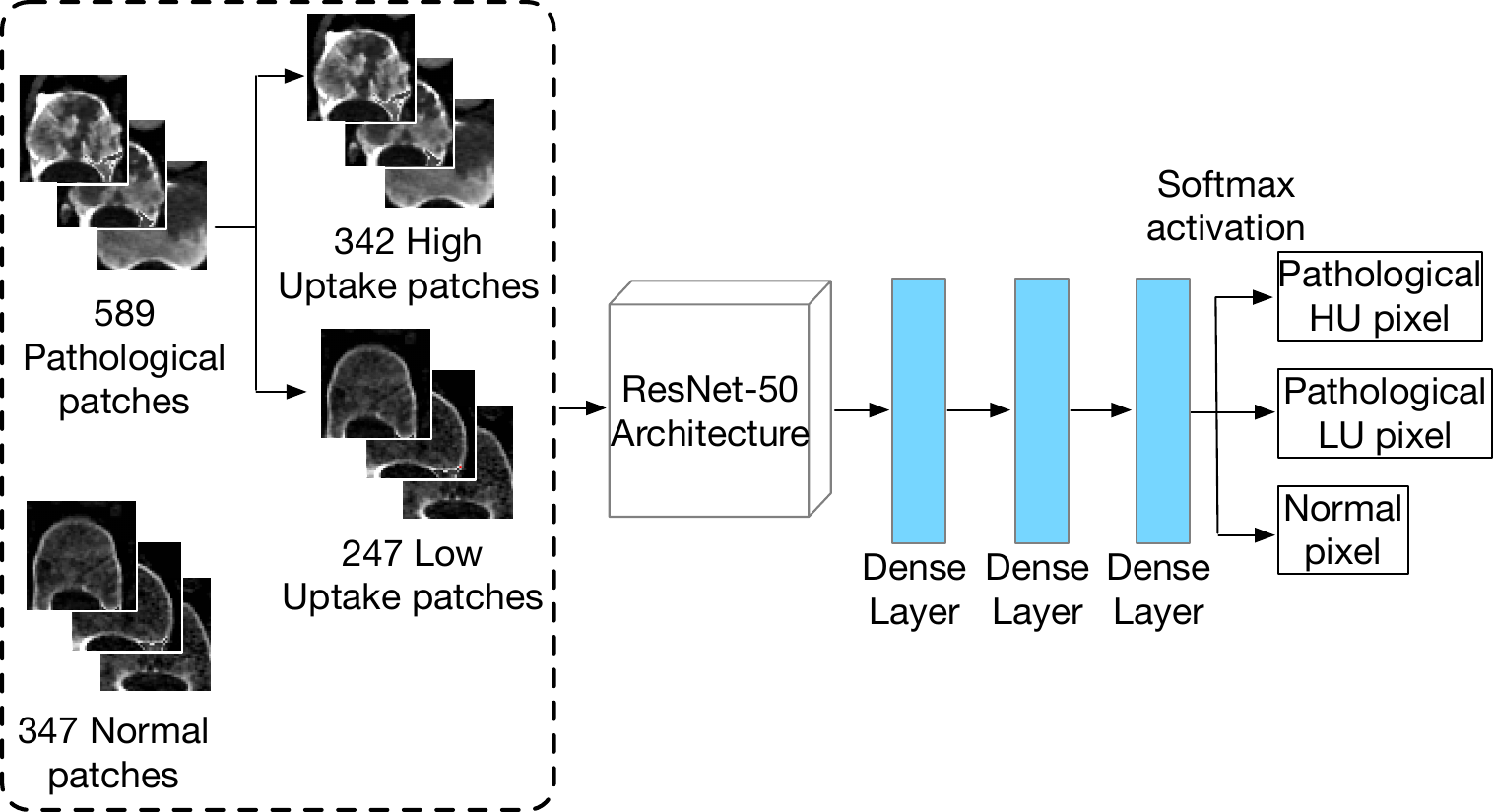}f
\caption{Overview of training procedure based on ResNet-50 architecture. Transfer learning method was utilized to extract features from pre-trained ResNet-50 model. }
\label{fig:resnet}
\end{figure}

To compare our framework to DNN's, we used pre-trained ResNet-50 architecture \cite{He}. We classified CT patches from our dataset into 3 classes ("Pathological HU", "Pathological LU", and "Normal"), using the ResNet-50 model, which was pre-trained on the ImageNet dataset \cite{Deng}. To fine-tune the ResNet-50 architecture three methods were examined. In the  first method,  all layers were frozen. In the second method, we unfroze the \newtext{last} 10 layers. In the first and second methods, after the last pre-trained layer  we added a global average pooling layer, a dropout layer, and an output layer activated by the softmax function to get the list of prediction scores. In the third method, we unfroze 2 last convolution blocks of ResNet-50 architecture for training, and then fed the extracted features into 2 fully connected dense layers with a rectified linear unit (ReLU) activation function and dropout regularization. To obtain the final 3 class probabilities we used a dense layer with the softmax activation function (see Fig. \ref{fig:resnet}). 

After classifying CT patches into 3 classes, in all three methods, we \newtext{changed} the “Pathological LU” obtained tags with the “Normal” tags\ynote{. We} then \ynote{} calculated the mean prediction of each vertebra. If at least one vertebra score was higher than a predefined cutoff, we classified the study as “Pathological”. We evaluated the ResNet-50 based framework using the 10-fold cross-validation technique. Table \ref{compare_table2} shows the AUC of the ResNet-50 based framework, for which the best option (third method above) reaches an AUC of 0.70.

\begin{figure}[h]
\centering 
\includegraphics[width=3in]{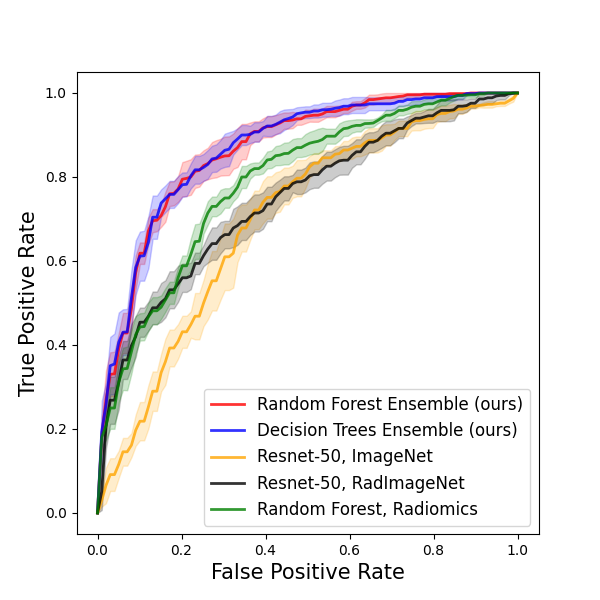}
\caption{Patients classification results evaluated using mean receiver operating characteristic curve (ROC) with variability.}
\label{fig:auc_comp2}
\end{figure}

Pretraining on ImageNet is most adequate for natural images and may not yield the best features for medical imaging. We thus utilized RadImageNet \cite{Mei} which is intended specifically to be used on radiological data. A pre-trained model with ResNet-50 architecture is used, in order to improve the performance of the DNN-based framework. We trained three DNNs, in the same way we trained ImageNet based models, the first one by freezing all layers,  the second one by unfreezing \newtext{last} 10 layers,  the third one by unfreezing 2 layers and adding 2 fully connected layers with  (ReLU) activation function and dropout regularization, then an output fully connected layer with the softmax activation function.   
We achieved AUC of 0.75 by freezing all layers, the results are shown in table \ref{compare_table2}.
In Fig. \ref{fig:auc_comp2} we show the receiver operating characteristic (ROC) curves for two our models (the best performance), the random forest model based on radiomics features (a good performance), the ResNet-50 model based on the RadImageNet pre-trained weights (a good performance), the ResNet-50 model based on the ImageNet pre-trained weights (a poor performance). 

\begin{figure}[h]
\centering 
\includegraphics[width=3in]{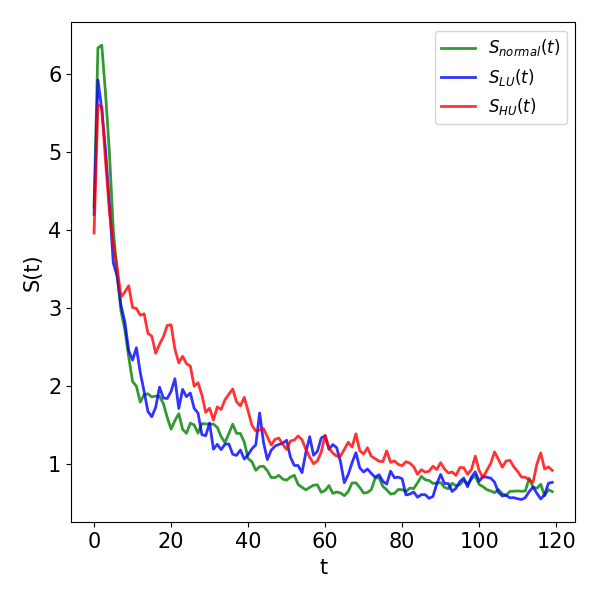}
\caption{Mean spectral response $S(t)$, Eq. \eqref{eq:S}, of "Normal", "Low Uptake" (LU), and "High Uptake" (HU) regions. HU regions have higher mean spectral response S(t) in almost every scales ($t \in [8,120]$), most notably in the range $[10,40]$. Pathological low-uptake has also a different mean spectral response, compared to normal patients.}
\label{fig:s1}
\end{figure}

In Fig. \ref{fig:s1} the mean spectrum (spectral response) $S(t)$ of patches' regions of interest is plotted.
Mean spectral response of high uptake regions are higher than the mean spectral response of normal regions in almost all scales ($t \in [8,120]$). This may explain the success of STV-based classification results. 
The code and the datasets are available at \href{https://technionmail-my.sharepoint.com/:f:/g/personal/srosnbrg_campus_technion_ac_il/EhT0LQd-RxFBuUq5MVLQD_0BwoWUbx8ztWT1SV0sd36ifg?e=E6sudd}{Project Repository}.

\section{Discussion and Conclusion}\label{sec:Conclusion}

We developed a novel classification model for differentiating pathological patients with skeletal metastases using CT scans, where PET information serves as ground truth in the training phase. Our model utilizes spectral TV signatures that were not used before in medical image classification, in general, and in bone metastases detection, in particular. 
In a standard procedure, radiologists diagnose skeletal metastases using PET/CT scans. Our approach opens ways to predict cases suspected of skeletal metastases based on CT only. This may enable early detection of metastases, as CT scanners are often more widely available. 
We investigated the ability to distinguish lesion bones from healthy bones. 
Our work demonstrates that spectral TV carries informative features for this task. 

This is a hard classification problem, as there is strong variability in the data and the CT cues are subtle.
We compared our results to radiomics features and to DNN based methods. Our method yields superior performance in terms of accuracy and AUC.
Spectral TV (STV) signatures \cite{hait} contain scale and space information that well differentiates objects of different contrast, size and structure. The lowly-correlated band characteristics allows to construct excellent ensembles that are resilient to noise and variability. In this study, we combined STV signatures into bands to form an ensemble of decision trees or random forests for classifying CT data. 
In ensemble learning, it is well known that most of the gain is obtained when the ensemble members are lowly correlated. Thus STV fits very well the ensemble framework, obtaining in each band different structural information. 
We believe additional classification tasks in CT and in medical imaging in general can benefit by the use of STV features.

\bibliographystyle{unsrt}
\bibliography{references}

\end{document}